\documentclass[10pt,twocolumn,letterpaper]{article}

\usepackage{ijcb}
\usepackage{times}
\usepackage{epsfig}
\usepackage{graphicx}
\usepackage{amsmath}
\usepackage{amssymb}

\ijcbfinalcopy 

\ifijcbfinal\fi

\begin{document}

\title{Detection of Fights in Videos: A Comparison Study of Anomaly Detection and Action Recognition}

\author{Weijun Tan, Jingfeng Liu\\
Jovision-Deepcam Research Institute\\
Shenzhen, China\\
{\tt\small {sz.twj@jovision.com,weijun.tan@deepcam.com}}\\
{\tt\small {sz.ljf@jovision.com,jingfeng.liu@deepcam.com}}
\and
{}
}

\maketitle

\begin{abstract}
   Detection of fights is an important surveillance application in videos. Most existing methods use supervised binary action recognition. Since frame-level annotations are very hard to get for anomaly detection, weakly supervised learning using multiple instance learning is widely used. This paper explores the detection of fights in videos as one special type of anomaly detection and as binary action recognition. We use the UBI-Fight and NTU-CCTV-Fight datasets for most of the study since they have frame-level annotations. We find that the anomaly detection has similar or even better performance than the action recognition. Furthermore, we study to use anomaly detection as a toolbox to generate training datasets for action recognition in an iterative way conditioned on the performance of the anomaly detection. Experiment results should show that we achieve state-of-the-art performance on three fight detection datasets. 

\end{abstract}

\section{Introduction}

Surveillance cameras are widely used in public places for safety purposes.  Enpowered by machine learning and artificial intelligence, surveillance cameras become smarter using automatic object or event detection and recognition.  Video anomaly detection is to identify the time and space of abnormal objects or events in videos.  Examples include industrial anomaly detection and security anomaly detection, and more. 

This paper studies a special type of anomaly detection - single-type abnormal event detection in videos.  Specifically we study the detection of fights in public places \cite{IPTA}, \cite{NTU_CCTV_Fight}, \cite{UBI_Fight}, \cite{8784746}, \cite{FightNet}, \cite{8078468}. But the algorithms can be easily extended to other single-type event detection, e.g., gun event detection.  A fight is a common event that needs the attention of safety personnel to prevent its escalation and becoming more destructive or even deadly.   

In video anomaly detection, more attention is given to the detection of general abnormal events.  It is not unfair since the algorithms for the detection of general abnormal events can mostly be used for the detection of a single-type abnormal event.  There are two major categories of anomaly detection algorithms.  The first algorithm is self supervised learning only using normal videos.  A model is learned to represent the normal videos.  Based on the modeling of the past frames, a prediction of a new frame is expected.  An abnormal event is reported if the actual new frame is too different from the predicted frame exceeding a pre-set threshold.  In this category, latent-space features or generative models can be used for both hand-crafted and deep learning features.  For more details of this algorithm, please refer to \cite{review}. 

The second is a weakly supervised learning approach, where a weak annotation label of if there is an abnormal event is presented in a video is provided.  In this case, typically, a multiple instance learning (MIL) is used \cite{UCF-Crime}.  From a pair of abnormal and normal videos, a positive bag of instances is formed on the abnormal video, and a negative bag of instances on the normal video.  The model tries to maximize the distance between the maximum scores of two bags.     

Anomaly detection is closely related to action recognition.  However, action recognition typically needs frame-level annotation, which is very hard to get for large-scale video datasets.  That is why the self and weakly supervised learning becomes prevailing.  For single-type abnormal event detection, the annotation of the frame-level labels become relatively easier.  For the fight detection, there are at least two datasets, the UBI-Fight \cite{UBI_Fight} and the NTU-CCTV-Fight \cite{NTU_CCTV_Fight}, which provide frame-level labels.  In this paper, we mostly use the UBI-Fight dataset in our study, while other datasets are used to benchmark our algorithms' performance.  With the frame-level labels, we can use supervised action detection.  It is fairly believed that supervised learning should outperform the self or weakly supervised learning.  However, this has not been well studied in the literature.

In this work, we do a comparison study of the weakly supervised learning vs. the supervised one.  Our goal is to give a suggestion on which one should be used in terms of performance, speed, and resource usage.  To our knowledge, except for \cite{UBI_Fight}, we are the first to explore using anomaly detection for the detection of single-type abnormal events in videos. 

Secondly, we study if we can use weakly supervised learning as a data generation tool to generate training data for the supervised action recognition.  The idea is to find the most reliable snippets out of the abnormal videos and the hardest snippets out of the normal videos.  The data generation approach helps the training of the supervised action recognition when an annotated dataset is not available.  Since we have frame-level annotation labels in the two datasets we use, we can compare the performance of this action recognition with the one using the frame annotation labels.   

\section{Related Work}

In this section we review literature on both the general anomaly detection and the fight detection in videos. 

\subsection{General Anomaly Detection}

Weakly supervised anomaly detection has shown substantially improved performance over the self supervised approaches by leveraging the available video-level annotations. These annotation only gives a binary label of abnormal or normal for a video. Sultani et al. \cite{UCF-Crime} propose the MIL framework using only video-level labels and introduce the large-scale anomaly detection dataset, UCF-Crime. This work inspires quite a few follow-up studies \cite{adgcn_cvpr19}, \cite{wsal_tip21}, \cite{UBI_Fight}, \cite{AAAI22}, \cite{CRFD}, \cite{IJCAI21}, \cite{MIST}, \cite{RTFM}, \cite{CRFD}. 
.

However, in the MIL-based methods, abnormal video labels are not easy to be used effectively. Typically, the classification score is used to tell if a snippet is abnormal or normal. This score is noisy in the positive bag, where a normal snippet can be mistakenly taken as the top abnormal event in an anomaly video. To deal with this problem, Zhong et al. \cite{adgcn_cvpr19} treat this problem as a binary classification under noisy label problem and use a graph convolution neural (GCN) network to clear the label noise. In RTFM \cite{RTFM}, a robust temporal feature magnitude (RTFM) is used to select the most reliable abnormal snippets from the abnormal videos and the normal videos.  They unify the representation learning and anomaly score learning by an temporal feature ranking loss, enabling better separation between normal and abnormal feature representations, improving the exploration of weak labels compared to previous MIL methods. In \cite{AAAI22} a multiple sequence learning (MSL) is used. The MSL uses a sequence of multiple instances as the optimization unit instead of one single instance in the MIL. In \cite{UBI_Fight} an iterative weak and self-supervised classification framework is proposed where their key idea is to add new date to the learning set. They use Bayesian classifiers to choose the most reliable segments for the weak and self supervised paradigms. 

There are very few work on the supervised learning for anomaly detection since the frame level annotation is very hard to get. Two examples are \cite{MMM19} and \cite{localizanomaly}.  

\subsection{Fight Detection Using Action Recognition}

Detection of fights in videos mostly follow the approach of action recognition. It is a simpler binary classification task to classify fight or non-fight actions.  Typical methods include 2D CNN feature extraction followed by some types of RNN, or 3D CNN feature extraction \cite{IPTA}, \cite{NTU_CCTV_Fight}, \cite{8784746}, \cite{FightNet}, \cite{8078468}.  

Early work uses the Hockey and Peliculas dataset \cite{Hockey}, and others, which are easy tasks. With pretrained 2D or 3D CNN feature extraction plus some feature aggregation techniques, the accuracy of the prediction can be very closed to 100\% \cite{FightNet}, \cite{8784746}, \cite{8078468}. Later a few more realistic datasets from surveillance or mobile cameras are made available, including the one in \cite{IPTA}, the NTU-CCTV-Fight dataset \cite{NTU_CCTV_Fight} and the UBI-Fight dataset \cite{UBI_Fight}. On the NTU-CCTV-Fights dataset, the frame mean average precision (mAP) is only 79.5\% \cite{NTU_CCTV_Fight}, and on the UBI-Fights dataset, the frame detection AUC is 81.9\% \cite{UBI_Fight}.       

\section{Proposed Methods}

In this section, we first compare supervised action recognition and weakly supervised anomaly detection in the detection of fights in videos.  Our purpose is not to present new action recognition or anomaly detection algorithms but rather to use existing algorithms to shed light on what directions we should go.  This serves as a baseline of how well we can do and motivates the solution we propose. 

\subsection{Action Recognition of Fights}

There are many approaches for action recognition on trimmed video clips, including 2D CNN+RNN, 3D CNN, with or without attention mechanism, transformers, and many variations.  For a review of action recognition, please refer to \cite{reviewAR}.  In this work, we use standard I3D CNN action recognition network \cite{I3D}, R(2+1)D CNN recognition network \cite{R2+1D}, and the ones with late temporal fusion using BERT \cite{BERT}, as shown in Fig 1(a).  This approach achieves the SOTA performance on the UCF101 \cite{UCF101} and HMDB51 \cite{JHMDB} action recognition datasets.   

In video action recognition tasks, the videos are already trimmed with the correct action boundary.  These videos can be short or long, with tens or thousands of frames.  One important point is how to select the right frames for the training and testing of the action recognition.  There are a few studies that propose to choose smartly which frames to use \cite{smartframe} or use all frames \cite{noleftbehind}.  In typical action recognition networks, 16, 32, or 64 frames are used in a video snippet sample.  Too many frames used in a step of training can easily cause out-of-memory problems since these frames need to be remembered in the GPU cache for the gradient calculation in the backpropagation.  When the video clip is too long, a typical method is to first cut off the clip into shorter and equal-length segments, then sample a snippet of the given length at a random location in the segment.  A problem with this sampling method is that the selected snippet may not be optimum for the action recognition training.  We will discuss this problem later in this work with experiment results. 

\subsection{Anomaly Detection of Fights}

To our knowledge, there is very little work that treats the detection of fights as an anomaly detection task except for \cite{UBI_Fight}.  This is probably because the annotated dataset is relatively easy to get, so people tend to use action recognition.  The second reason is that people do not realize the power of anomaly detection when the action recognition methods are available to use off-shelve.  In reality, when we try to deploy such a system into a practical application, we usually find that there are always cases not covered by current datasets.  In this case, the weak supervision method makes the work of preparing the new dataset a lot easier than that for the supervised action recognition.       

We have reviewed the MIL-based anomaly detection methods in the previous section.  We find that the RTMF \cite{RTFM}, shown in Fig. 1(b) gives great performance using a relatively easy architecture.  The key innovation is to use a temporal feature magnitude to discriminate abnormal snippets from normal snippets. The AUC of the RTMF on the UCF-Crime dataset is 84.03\%, when only the RGB modal is used with the I3D backbone. 

In this work, we modify the RTMF framework \cite{RTFM} as our anomaly detector for fight detection in videos. In addition to the basic binary cross entropy loss, the temporal smoothness loss and sparsity loss \cite{UCF-Crime}, a new loss on the feature magnitude on the top-k selected abnormal and normal segments is introduced,  
\begin{equation}
    loss_{FM} = 
        \begin{cases}
            max(0, m-d_{k}(X_i,X_j)), \, if\, y_i=1,\,y_j=0\\
            0, otherwise \,\\
        \end{cases}
    \label{eq1}
\end{equation}
where $y$ is the label, $X$ is a snippet, $d_{k}$ is a distance function that separates the top-k magnitude segment features of the abnormal video $y_i$ and the normal video $y_j$, and $m$ is a predefined margin. The form of this loss function is similar to the MIL ranking loss in \cite{UCF-Crime}, but it is now on the feature magnitude and on the top-k magnitude segments. 

\subsection{Iterative Anomaly Detection and Action Recognition}

Other than the extensive comparative study of anomaly detection and action recognition, we propose to use anomaly detection and action recognition in an iterative manner, as shown in Fig. 1.  Our novelty is to use anomaly detection to find good-quality training data for the action recognition.  Since the UBI-Fight dataset \cite{UBI_Fight} has the frame-level annotation, we use it extensively in our study.  When anomaly detection is used, we only use the video-level annotation and ignore (or assume unknown) the frame-level annotations.  We use the frame-level annotation in the supervised action recognition. 

In anomaly detection dataset, in every abnormal positive video, we know there are at least one or more abnormal snippets.  In every normal negative video, we know all snippets are negative.  So intuitive thinking is that we only need to generate positive training data for action recognition.  In our study, we find that the negative training data can also be refined.  When random data sampling is used, all samples have an equal chance of being picked for a training epoch.  However, not all samples contribute equally to the training process.  To increase the model's discrimination capability, hard negative samples are preferred.  They help not only the performance but also the convergence. 

In the RTFM training process, every video is divided into 32 equal-length segments, the same as what is done in \cite{UCF-Crime}.  The features of all 16-frame snippets in a segment are averaged as the feature for this segment.  The top three segments whose feature magnitude is the largest are picked, and their average magnitude is used in training.  At the inference time, an original video snippet is used, and a classification score is generated for every snippet.  This score is used to pick the snippets for the action recognition.  Please note that this is different from using the classification score in \cite{UCF-Crime}, where the score is used to pick the most probable segment in training.  In RTFM, the score is used at the inference time after the training is already done. 

For an abnormal positive sample, this score is typically large, very close to 1.0.  So we use a threshold like 0.995 to pick snippets whose scores are higher than it.  We notice that when RTFM converges too fast or overfits, the picked snippets may have some wrong results, meaning that normal snippets are picked as abnormal snippets, which is very detrimental to the action recognition performance. 

For normal negative samples, their scores are typically small but can be large.  Our strategy is to use a threshold around 0.5.  The samples whose scores are larger than this threshold are hard samples.  One other factor we need to consider is to balance the number of generated positive and negative samples.  We can adjust this threshold to make them close to each other.     

There are two cases in which this iterative method can be used.  In anomaly detection, we use the I3D backbone \cite{I3D} pre-trained on the Imagenet \cite{imagenet} and Kinetics-400 \cite{Kinetics400} datasets.  We use the above method to generate training samples for the action recognition network.  In the first case, the backbones of the anomaly detection and the action recognition are the same.  In this case, after the first iteration of the action recognition, its updated backbone can be used to regenerate the video features for the anomaly detection in the second iteration. 

In the second case, the backbone of the action recognition network is different from that of the anomaly detection network.  Then after the anomaly detection networks generate training samples for the action recognition, this process can stop.  Even though the action recognition network can regenerate video features for the anomaly detection, then the anomaly detection becomes a different network.  This process is not preferred and is not tested in our work.  In this sense, the anomaly detection network is more like a toolbox for generating training data for a different action recognition network.   

\section{Experiments}

\subsection{Datasets}

A small fight detection dataset in \cite{IPTA} is used. It consists of trimmed video clips good for action recognition. It includes 150 fight videos and 150 normal videos. 

\textbf{UBI-Fight}. The UBI-Fights dataset is released in \cite{UBI_Fight}. It provides a wide diversity in fighting scenarios. It includes 80 hours of video fully annotated at the frame level consisting of 1000 videos, where 216 videos contain a fight event, and 784 are normal daily life situations. 

\textbf{NTU-CCTV-Fight}. The NTU-CCTV-Fight is released in \cite{NTU_CCTV_Fight} . It includes 1000 videos, each one of which includes at least one fight event. No normal videos are included. Frame-level annotations are provided in a JSON format. It mixes two types of cameras - surveillance cameras and mobile cameras, making fight detection hard to detect in this dataset.  

\subsection{Implementation Details}

For anomaly detection, we use the RTMF codebase \cite{RTFM} in PyTorch. We use the I3D with Resnet18 backbone \cite{I3D} for the video feature generation. A training rate of 1E-3 is used, and the training runs 100 epochs. Two dataset iterators, one for the abnormal data and the other for the normal data, are used. This way, the pairing of abnormal and normal data is random, even when the numbers of abnormal and normal samples are different. 

For the action recognition, we also use a codebase in PyTorch. We use the I3D network with Resnet18 backbone \cite{I3D}, and the R(2+1)D network with Resnet34 backbone \cite{R2+1D}. We use the BERT similar to \cite{BERT}. For the I3D network, we use the model pre-trained on the Imagenet \cite{imagenet} and Kinetics-400 \cite{Kinetics400} datasets . For the R(2+1)D network, we use the model pre-trained on the IG65 dataset \cite{IG65}. Our start learning rate is 1E-5, and the learning rate reduces by a 0.1 factor on a plateau with 5-epochs patience. The checkpoint with the best validation accuracy is saved for evaluation on the test dataset. 

\textbf{Evaluation Metric}. For action recognition, we use accuracy as the metric. For all the results on the UBI-Fight dataset, \cite{UBI_Fight}, the frame AUC is used. While for the results on the NTU-CCTV-Fight dataset \cite{NTU_CCTV_Fight}, the frame mAP is used. 

\subsection{Action Recognition Results}

We first show the action recognition results in accuracy (Acc, 4th column), frame AUC (5th column), and mAP (6th column) on a few fight detection datasets. These results all outperform previous SOTA results. They also serve as a baseline or bound for our iterative anomaly detection and action recognition approach. 

\begin{table*}[tb]
\begin{center}
  \begin{tabular}{|l|c|c|c|c|l|}
    \hline
    Dataset  & Method & Frames & Acc & AUC & mAP   \\
    \hline
    Dataset \cite{IPTA} & 2DCNN+Bi-LSTM & 5 random & 0.72 &  - & -\\
    Dataset \cite{IPTA} & ours R(2+1)D-BERT & 32 random & \textbf{0.9562} &  - & -\\
    \hline
    UBI-Fight\cite{UBI_Fight} & Iterative WS/SS$^{[1]}$ & 16 avg & - &  0.819 & - \\
    UBI-Fight & ours R(2+1)D-BERT & 32 random & 0.9177 & \textbf{0.9150} &  - \\
    UBI-Fight & ours I3D & 32 random & 0.8861 & 0.9058 &  - \\
    UBI-Fight & ours I3D & 16 random$^{[2]}$ & 0.80 & 0.8149 &  - \\
    
    \hline
    NTU-CCTV-Fight\cite{NTU_CCTV_Fight} & RGB+Flow-2DCNN & 16 & - &  - & 0.795 \\
    NTU-CCTV-Fight & ours R(2+1)D-BERT & 32 random & 0.8244 & 0.8715 & \textbf{0.8275} \\    
    
  \hline

  \end{tabular}
  \caption{Action recognition results on three fight detection datasets. Notes [1]: This method uses weakly supervised anomaly detection. We put it here for comparison with our action recognition results. [2]: We use multiple sets of snippets and average the features as input to the classification layers for better performance.}
  \label{Table2}
  \end{center}
  \end{table*}

Shown in Table \ref{Table2} are the results where we test both the standard I3D \cite{I3D}, and the more advanced R(2+1)D-BERT \cite{BERT}. Other than the backbone and classification network, the number of frames input to the 3D CNN and how they are sampled are also important. For long action recognition samples, an ideal sampling method is to use all available frames in every epoch of the training, similar to what is done in anomaly detection. However, due to the limited memory in GPU, it is very hard to do so if the backbone network is trained. In our test, we find that I3D with 16 frames does not have good performance and realize that the number of fewer frames may be one of the causes. So we sample multiple sets (typically 4 or 8)  of 16 frames in an epoch and average the features before it is fed to the classification layers. The accuracy result improves from 0.778 to 0.800 in Table \ref{Table2}.      

In this subsection, we focus on the anomaly detection results on the UBI-Fight dataset \cite{UBI_Fight}. We use the RTFM \cite{RTFM} as the anomaly detection method. All videos are first divided into 32 equal-length segments. Then features are extracted for all 16-frame snippets sequentially in every segment, and these features are averaged as the representing feature for this segment. The standard I3D pre-trained on the Imagenet is used. The dimension of the feature is 1024.  

\begin{table}[tb]
\begin{center}
  \begin{tabular}{|l|c|l|}
    \hline
    Method & Crop &  AUC  \\
    \hline
    I3D16F Action Recognition (AR) & 1-crop & 0.8149 \\  
    \hline
    RTFM  & 10-crop & 0.9192  \\
    RTFM  & 1-crop & 0.9097  \\
    RTFM w/o MSNL & 1-crop & 0.8692  \\
    \hline
    RTFM-Action Recognition (AR) & 1-crop & 0.9129  \\
    RTFM-AR w/o MSNL & 1-crop & 0.8708 \\
    RTFM-AR w/o MSNL, 1-snippet & 1-crop & 0.8408 \\

  \hline

  \end{tabular}
  \caption{Anomaly detection results vs. action recognition results on the UBI-Fight dataset. AR means action recognition.}
  \label{Table3}
  \end{center}
  \end{table}

Shown in Table \ref{Table3} are the anomaly detection results with their corresponding action recognition results for reference. Since all snippets use 16 frames, we put the I3D with 16-frames (I3D16F) in the table for comparison. The original RTFM uses 10-crop data augmentation. However, in practice, particularly at inference time, it is impossible to use a 10-crop, so we use the 1-crop result as our baseline, with an AUC=0.9097. We see that this value is a lot better than the I3D16F result for a few reasons. First, the RTFM uses a powerful MS-NL aggregation neck before the classification head. When we turn off this aggregation neck, the AUC becomes 0.8692, which is still better than the I3D16F's 0.8149.

We do more analysis and find that this is contributed to using all 16-frame snippets in the segment. As explained in Subsection 3.1, in every epoch of the action recognition training, due to the limit of the GPU memory, only one set of 16 frames is randomly sampled. We do an experiment in Subsection 4.1, where we use multiple sets of 16 frames and achieve better performance. This effect becomes less noticeable when 32 or 64-frame snippets are used. In anomaly detection, since the backbone is not trained, so the features of all snippets can be calculated and used in the training of the classification head. 

The second cause is the nature of the MIL in the anomaly detection \cite{UCF-Crime}. The MIL uses the segments whose scores or the feature magnitudes in the RTFM are the maximum instead of randomly chosen ones. Particularly on the negative samples, this has an effect of using the hard samples. We do a test to confirm this effect, where we randomly choose a negative segment, and the performance of the anomaly detection gets worse.    

\subsection{Anomaly Detection or Action Recognition?}

\textbf{Devised supervised anomaly detection.} We devise a set of supervised anomaly detection experiments to do our analysis. Instead of using the whole abnormal videos as positive samples in the RTFM, we use the frame-level annotation and extract the ground truth positive portions as the positive samples. These are the same as the trimmed positive samples used in the action recognition in Subsection 4.1. There are no changes in the negative samples. We remove all other loss functions and only use the BCE-loss function. We also remove choosing the top-3 feature magnitude segments and replace them with a random choice. Since all the positive segments use ground truth abnormal portions, it does not choose normal snippets as abnormal ones. This essentially makes the RTFM a different implementation of action recognition. There are still two differences. The first one is that the backbone is not trained. The second one is that all snippets of a segment are still used for feature calculation. The results are shown in Table \ref{Table3} denoted by RTFM-AR. The AUCs are now a little bit higher than the ones using weak supervision. The AUC for the one without MSNL is still higher than the I3D16F whose AUC is 0.8149. Going one step further, we remove the feature averaging on all snippets in a segment and use one randomly chosen snippet per segment instead. At the same time, we reduce the number of segments per video and change other settings so that the effective batch number is equivalent to that in the action recognition. The AUC reduces significantly to 0.8408, but still higher than that of the I3D16F. This is very likely because, in the RTMF action recognition, the training runs a lot faster since the I3D backbone is not trained. As more epochs are run, all the snippets have a better chance of being picked. While in the standard I3D16F, since the backbone is trained, the training runs very slowly, and some snippets may never get a chance to be picked.  

From these experiments, we notice that with same backbone and similar classification head, the weakly supervised anomaly detection (RTFM w/o MSNL) and supervised action recognition (RTFM-AR w/o MSNL) can achieve almost the same performance. With additional MSNL neck, the anomaly detection can achieve better performance. Since the data preparation and training of the anomaly detection are much easier, we conclude that the anomaly detection is preferred 
for the detection of a single-type abnormal event in video.   

\subsection{Iterative Anomaly Detection and Action Recognition}

\begin{figure}[t]
    \centering
    \includegraphics[scale=0.6]{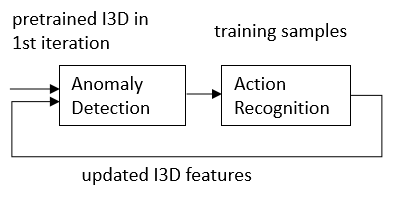}
    \caption{pipeline of iterative anomaly detection and action recognition.} 
    \label{fig2}
\end{figure}

In this subsection, we explore the iterative anomaly detection and action recognition, as shown in Fig.\ref{fig2}. The primary purpose is to find out if anomaly detection can be used as an effective toolbox to generate training samples for standard action recognition, and to achieve the possible best performance of a supervised action recognition.  

In the first iteration, we first train the RTFM \cite{RTFM} anomaly detector using the features extracted from an I3D network pre-trained on the Imagenet. In order to have the best quality snippet selection for the action recognition, we keep the MSNL aggregation neck in the RTFM. After the training is done, we run the training data in the RTFM inference mode. Snippets from the abnormal videos whose scores are larger than 0.995 are selected as positive samples, and snippets from the normal videos whose scores are larger than 0.5 are selected as hard negative samples. Then we train the standard I3D16F action recognition network using these generated training samples. Please note that the validation and test data do not go through the anomaly detection generation. This completes the first iteration of the anomaly detection and action recognition interaction. 

In the second iteration, the updated I3D backbone is used to generate a new dataset for anomaly detection. And the same training process for the RTFM follows on the newly generated data. Similarly are the new training data generation for the action recognition and the new action recognition training. This process can keep going for a few iterations. 

\begin{table}[tb]
\begin{center}
  \begin{tabular}{|l|c|c|l|}
    \hline
    Dataset & Iter & RTFM &  I3D-AR \\
    \hline
    UBI-Fight & 0-th  & - & 0.8149 (AUC) \\  
    UBI-Fight & 1st  & 0.9097 & \textbf{0.8789} \\
    UBI-Fight & 2nd  & 0.9279 & 0.8115 \\  
    \hline
    NTU-CCTV-Fight & 0-th  & - & \textbf{0.8275}(mAP) \\  
    NTU-CCTV-Fight & 1st  & 0.8140 & 0.8070 \\
  \hline
  \end{tabular}
  \caption{Iterative anomaly detection and action recognition on the UBI-Fight dataset. AR means action recognition.}
  \label{Table4}
  \end{center}
  \end{table}

Shown in Table \ref{Table4} are the anomaly detection and action recognition AUCs in the first two iterations.  We call the previous standard action recognition the 0-th iteration.  It is observed that the I3D16F AUC improves from 0.8149 to 0.8789 in the first iteration.  This proves that the RTFM anomaly detector can be used to generate training samples for the I3D16F action recognition.  We do some deeper analysis and find that the hard negative samples contribute to this performance improvement.  If the I3D16F action recognition can be trained well enough and as many as possible snippets can be used, we believe that the I3D16F itself can also achieve this AUC of 0.8789. 

However, the AUC does not improve further in the 2nd iteration.  Analysis shows that even though the RTFM AUC improves a little bit from the 1st iteration, the scores' values also get larger.  In this sense, the RTFM is overfitted, and more normal video snippets have scores close to 1; therefore are selected erroneously as positive samples for the action recognition.  That is why the I3D16F action recognition AUC goes back to 0.8815 in the 2nd iteration.   

We repeat this same experiment on the NTU-CCTV-Fight dataset but with I3D32F since 32-frames give better performance than 16-frames.  Since there are surveillance cameras and mobile cameras mixed in this dataset, the fights are harder to detect in this dataset.  In the first iteration, The anomaly detection's AUC is 0.8140, and the I3D32F action recognition's mAP is 0.8070, worse than that in the 0-th iteration, which is 0.8275.  Our analysis shows that the performance of the anomaly detection must be good enough to generate good quality training samples for the action recognition.  In Table \ref{Table3}, the anomaly detection AUC is 0.9097, while it is only 0.8140 in \ref{Table4}. We conjecture that there exists a threshold between 0.8140 and 0.9097.  When the anomaly detection's AUC is larger than the threshold, the generated training samples will help the performance of the subsequent action recognition. Since the performance does not improve in the 1st iteration, we do not continue the 2nd iteration. But overall, the best action recognition performance (0.8789 AUC on UBI-Fight) is about the same as that of the anomaly detection (0.8692 RTFM w/o MSNL on UBI-Fight in Table 2).  

\subsection{Comparison with SOTA Results}
Our results in Table \ref{Table2} are better than existing SOTA results on the three datasets.  Furthermore, if we take the best result out of all our studies, the best AUC on the UBI-Fight dataset \cite{UBI_Fight} is 0.9192 in Table 2.  

\section{Conclusion}

In this paper, we do a comparison study of the weakly supervised anomaly detection and the supervised action recognition.  Based on our experiment results, we find that anomaly detection can work as well as or even better than action recognition.  Since the weak supervision annotations are a lot easier to get, it makes anomaly detection a very good choice for the detection of single-type abnormal events in videos. 

We also find that anomaly detection can be used as a toolbox for the generation of training data for supervised action recognition in some conditions.  In this study, the condition is that the anomaly detection's AUC in the first iteration must be good enough.  This may not generalize to other different scenarios.  

{\small
\bibliographystyle{ieee}
\bibliography{egbib}
}

\end{document}